\documentclass[conference]{IEEEtran}
\IEEEoverridecommandlockouts
\usepackage{cite}
\usepackage{amsmath,amssymb,amsfonts}
\usepackage{algorithm}
\usepackage{algpseudocode}
\usepackage{graphicx}
\usepackage{textcomp}
\usepackage{xcolor}
\usepackage{multirow}
\usepackage{booktabs}
\usepackage{subcaption}
\usepackage{array}

\def\BibTeX{{\rm B\kern-.05em{\sc i\kern-.025em b}\kern-.08em
    T\kern-.1667em\lower.7ex\hbox{E}\kern-.125emX}}

\DeclareRobustCommand*{\IEEEauthorrefmark}[1]{%
    \raisebox{0pt}[0pt][0pt]{\textsuperscript{\footnotesize\ensuremath{#1}}}}

\begin{document}

\title{LCE: A Framework for Explainability of DNNs for Ultrasound Image Based on Concept Discovery
\thanks{
*Xun Gong is the corresponding author (xgong@swjtu.edu.cn).

This study is partially supported by National Natural Science Foundation of China (62376231), Sichuan Science and Technology Program (24NSFSC1070), Science and Technology Project of Sichuan Provincial Health Commission (23LCYJ022), Fundamental Research Funds for the Central Universities (2682023ZDPY001, 2682022KJ045).

This study adhered to the principles of the Helsinki Declaration and obtained informed consent (SWJTU-2208-SCS-072). Approval was obtained from the Clinical Research Ethics Committee of the Affiliated Hospital of Southwest Jiaotong University (Chengdu Third People's Hospital).
}
}

\author{
    \IEEEauthorblockN{Weiji Kong\IEEEauthorrefmark{1,2,3}, Xun Gong\IEEEauthorrefmark{1,2,3,*}, Juan Wang\IEEEauthorrefmark{4}}
    \IEEEauthorblockA{\IEEEauthorrefmark{1}School of Computing and Artificial Intelligence, Southwest Jiaotong University, Chengdu, China}
    \IEEEauthorblockA{\IEEEauthorrefmark{2}Engineering Research Center of Sustainable Urban Intelligent Transportation, Ministry of Education, Chengdu, China}
    \IEEEauthorblockA{\IEEEauthorrefmark{3}Manufacturing Industry Chains Collaboration and Information Support Technology Key Laboratory of Sichuan Province, \\ Southwest Jiaotong University,
Chengdu, China}
    \IEEEauthorblockA{\IEEEauthorrefmark{4}Department of Ultrasound, The Third People’s Hospital of Chengdu, \\Affiliated Hospital of Southwest Jiaotong University, Chengdu, China}
}

\maketitle

\begin{abstract}
Explaining the decisions of Deep Neural Networks (DNNs) for medical images has become increasingly important. Existing attribution methods have difficulty explaining the meaning of pixels while existing concept-based methods are limited by additional annotations or specific model structures that are difficult to apply to ultrasound images. In this paper, we propose the Lesion Concept Explainer (LCE) framework, which combines attribution methods with concept-based methods. We introduce the Segment Anything Model (SAM), fine-tuned on a large number of medical images, for concept discovery to enable a meaningful explanation of ultrasound image DNNs. The proposed framework is evaluated in terms of both faithfulness and understandability.  We point out deficiencies in the popular faithfulness evaluation metrics and propose a new evaluation metric. Our evaluation of public and private breast ultrasound datasets (BUSI and FG-US-B) shows that LCE performs well compared to commonly-used explainability methods. Finally, we also validate that LCE can consistently provide reliable explanations for more meaningful fine-grained diagnostic tasks in breast ultrasound.
\end{abstract}

\section{Introduction}
In recent years, Deep Neural Networks (DNNs) model have become mainstream in medical image analysis, outperforming even human doctors on various tasks in different medical image modalities\cite{chen2022recent}. As a black-box model, DNNs have difficulty explaining what happens to the input image within the complex structure of DNNs. Lack of explainability leads to difficulty in gaining the trust of doctors or patients, making it challenging to put medical DNNs into clinical use, even if they perform well. In addition, some laws and regulations \cite{kaminski2021right,kop2021eu} restrict the use of AI in human society.

Explainable Artificial Intelligence (XAI) has increasingly come to the fore as a way of unlocking the black-box. For computer vision tasks, XAI tends to explain the output of the model by identifying what the most important pixels are. There are already a large number of XAI methods that can provide a variety of explanations, such as attribution methods and concept-based methods.

Attribution methods can conveniently improve the explainability of medical DNNs \cite{zhao2018respond,zhu2019guideline,li2020efficient}. However, these methods are prone to false activations. They can only explain where in the image a pixel is located that is crucial for decision-making but are unable to identify which visual features drive decision-making at these locations. More importantly, explanations that do not match real-world meanings do not inspire confidence in doctors or patients.

\begin{figure*}[h]
  \centering
  \includegraphics[width=1.0\textwidth]{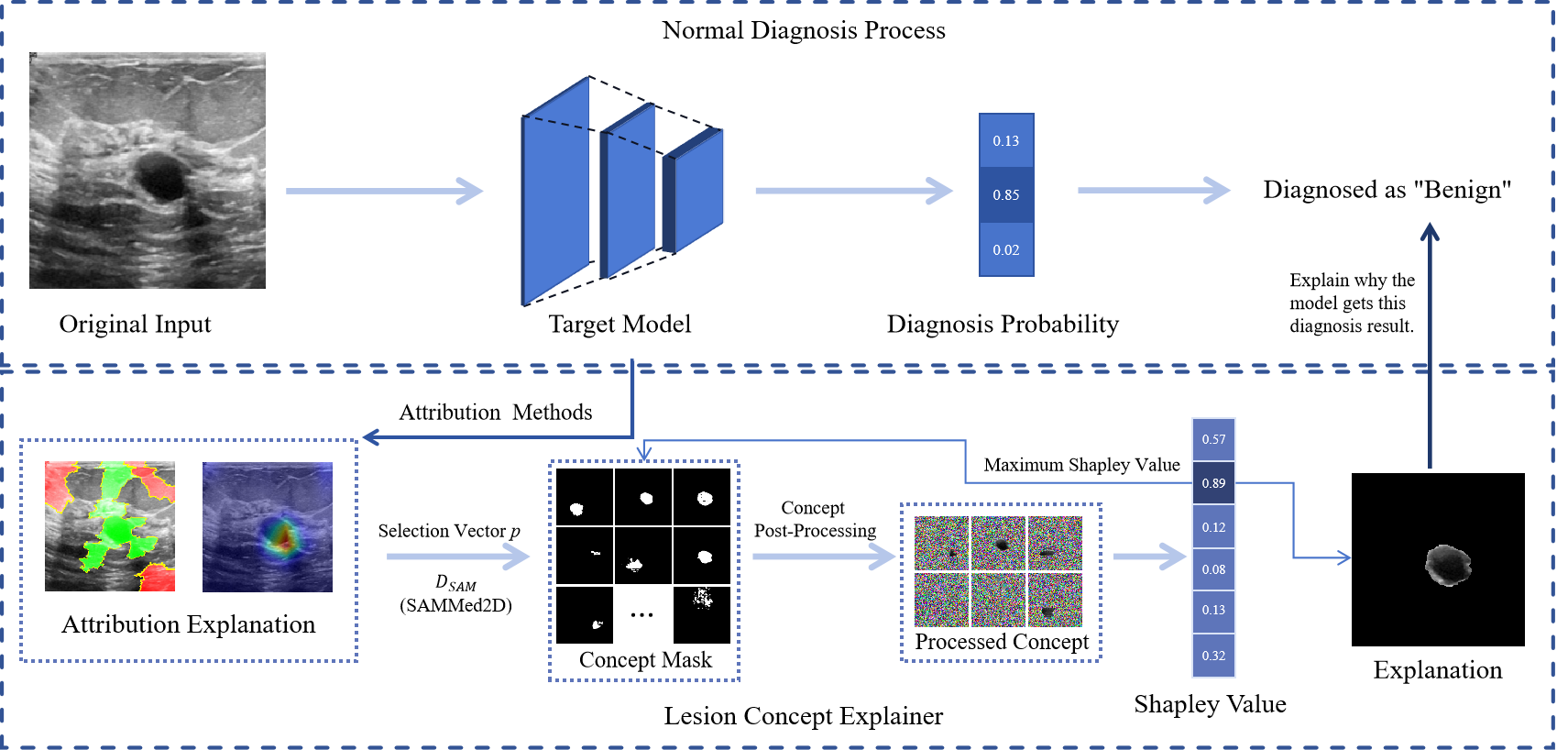}
  \caption{The Lesion Concept Explainer (LCE) Framework. For any black-box DNN model, attribution methods are first used to generate explanations, which are then used to guide $D_{SAM}$ to explore and obtain concept masks. The concept masks are then post-processed. Finally, the Shapley value is applied to identify the explanations that are most crucial for model's decisions.}
  \label{fig:method}
\end{figure*}

Concept-based methods \cite{bau2017network,kim2018interpretability} can provide more meaningful explanations, but these techniques usually require additional provision of manually predefined concepts. This is difficult to implement in the medical field, especially for ultrasound images, which are fast and low-cost, and adding several times the cost for explanation is like putting the cart before the horse. There are also some studies \cite{uehara2019prototype,gallee2023interpretable} that attempt to use prototypes to bring explainability to medical DNNs, but these require a specific model structure, and prototypes derived from low-cost ultrasound images may be less accurate.

Explain any concept (EAC)\cite{sun2024explain} pioneered the combination of the SAM\cite{kirillov2023segment}, which has excellent zero-shot segmentation capability, with Shapley value\cite{shapley1953value}, for efficient explanation. Concept discovery is the most fundamental step in EAC. However, it is difficult for EAC to discover medically meaningful concepts from medical images due to the huge gap between normal images and medical images.

Inspired by EAC\cite{sun2024explain}, in this paper, we propose a framework named Lesion Concept Explainer (LCE), which uses attribution methods to guide existing SAM fine-tuned on a large amount of medical image data in combination with Shapley values for concept discovery and explanation. We implement LCE on the public breast ultrasound dataset BUSI\cite{al2020dataset} and the private breast ultrasound dataset FG-US-B using ResNet50\cite{he2016deep} as the target model.

The faithfulness metrics $Insertion$ and $Deletion$\cite{petsiuk2018rise} can reflect the consistency of the explanation with the decision process within the model. Since the effect of the explanation size is not taken into account, the original image is assumed to be the best explanation, even if it does not make sense, when evaluated by $Insertion$ and $Deletion$. To solve the above problem, we add the explanation size as a weight in $Insertion$ and $Deletion$ and propose a new evaluation metric, $Effect\ Score$. Intuitively, an explanation is given to a person, and an understandable explanation is a good one. We combine the objective faithfulness evaluation of $Effect\ Score$ and the human-centered evaluation of understandability to evaluate LCE and other popular explainability methods. The result shows that LCE outperforms most popular methods in terms of faithfulness.  In the evaluation of understandability, LCE is preferred by professional sonographers.

Finally, we explored the performance of LCE in the more meaningful fine-grained diagnosis of breast ultrasound. The results show that LCE consistently yields reasonable explanations regardless of fine-grained level. Notably, LCE is model-agnostic and can provide explanations for any diagnostic model of medical images, and the whole framework requires no additional cost, making it ideal for low-cost ultrasound images.

The contributions of this paper are summarized as follows: 
\begin{itemize}
    \item We present a framework for explainability that enables understandable explanations of DNNs trained on low-cost ultrasound images.
    \item We integrate the attribution method with the concept-based method, synthesizing the strengths of both methods and improving the quality of explanations.
    \item We identify the shortcomings of popular faithfulness evaluation metrics and make adjustments accordingly.
    \item We validate the stability of applying our framework in a fine-grained diagnostic task in breast ultrasound.
\end{itemize}

\section{Related Work}
\subsection{Segment Anything Model (SAM)}

In recent years, the development of AI has focused on large-scale foundation models, resulting in a number of exceptional products\cite{bommasani2021opportunities,dosovitskiy2020image,kirillov2023segment}. These models, with billions or more parameters, exhibit robust generalization and problem-solving potential after extensive training on large datasets, making them versatile for various downstream tasks.
Among them, Segment Anything Model (SAM)\cite{kirillov2023segment} has caused a sensation with its powerful zero-shot segmentation performance. More importantly, SAM has learned general concepts about objects, including those not seen during training.

However, the direct application of SAM to the ultrasound image segmentation task is not ideal due to the significant differences between normal images and medical images \cite{mazurowski2023segment}. To address this problem, some studies have proposed methods to improve SAM and make it applicable to the field of medical image segmentation \cite{cheng2023sam,ma2024segment}. Of these, MedSAM \cite{ma2024segment} has fine-tuned SAM using a medical image dataset containing more than one million medical images with different diseases and modalities, and its excellent performance in medical image segmentation has been verified through extensive experiments. Similarly, SAMMed2D\cite{cheng2023sam} used the largest medical image segmentation dataset to date and introduced Adapter \cite{houlsby2019parameter} to fine-tune SAM. In this paper, we use existing SAM fine-tuned on medical images to explore the ability of SAM to discover concepts in medical images without additional training costs.

\subsection{Attribution methods}
Attribution methods primarily use techniques such as heat maps\cite{zhou2016learning} or saliency maps \cite{simonyan2013deep} to identify key regions that provide visual cues for model classification.  Class Activation Mapping (CAM) and its derivative methods \cite{zhou2016learning,selvaraju2017grad}can generate heat maps to show the contribution of each pixel in an image to the classification. CAMs are often used to explain medical image models, for example,\cite{irvin2019chexpert} used Grad-CAM\cite{selvaraju2017grad} to analyze chest X-ray images and obtained clear visualizations of pleural effusions, and \cite{izadyyazdanabadi2018weakly} used Multi-Layer CAM to localize Confocal Laser Endomicroscopy Glioma images.

In addition to methods such as CAM, which obtains significance scores from within the model, Local Interpretable Model-agnostic Explanations (LIME)\cite{zhou2016learning} observes changes in model output by simplifying complex models and perturbing input data for local superpixel explanations. \cite{petsiuk2018rise} used a randomly sampled mask to obtain pixel contributions. \cite{shrikumar2017learning} introduced DeepLift to assign contribution scores by comparing neuron activations. \cite{sundararajan2017axiomatic} proposed IntGrad, which combines invariance and sensitivity. \cite{lundberg2017unified} incorporated Shapley value\cite{shapley1953value} into LIME and DeepLift to form a unified framework for explanation, SHAP.  \cite{zhu2019guideline} integrated perturbation operations similar to LIME and SHAP to explain important features in the diagnosis of CT lung images. Similarly, \cite{bhandari2023exploring} used LIME and SHAP to explain a diagnostic model for renal abnormalities. However, the explanations given by these attribution methods are not always understood by medical image specialists.

\subsection{Concept-based methods}
In contrast to attribution methods, concept-based methods prefer explanations that are consistent with real-world concepts. \cite{bau2017network} proposed a framework to quantify the explainability of DNNs by evaluating the consistency between hidden units and semantic concepts. \cite{kim2018interpretability} introduced the Test of Concept Activation Vector (TCAV) to measure the sensitivity of the model to pre-defined concepts. \cite{clough2019global} applied the TCAV to the detection of cardiac disease in cine-MRI using clinically known biomarkers as concepts. However, the pre-definition of a concept set is challenging in the field of medical image processing as it requires the intervention of experts.

Some concept-based methods, which do not require human intervention, can automatically extract concepts from images. Prototype-based methods\cite{chen2019looks} can guide the network to use representative concepts for decision-making. \cite{uehara2019prototype} used prototypes to explain histological image classification. \cite{gallee2023interpretable} combined prototype learning with privileged information in thoracic CT diagnosis, balancing explainability and accuracy. \cite{sun2024explain} proposed Explain Any Concept (EAC), which uses zero-shot segmentation of SAM to discover concepts. Nevertheless, applying these methods to ultrasound images is challenging due to their low-cost and limited information compared to CT and MRI.

\begin{algorithm}[t]
%\caption{Lesion concept discovery$(I, M, D_{SAM}, G_l, G_f, q, k)$\label{al1}}
	\caption{Lesion concept discovery\label{al1}}
	\begin{algorithmic}[1]
		
		\Require Ultrasound Image $I$, Target Model $M$, Concept Discoverer $D_{SAM}$, Concept Guide $G_l$ and $G_g$, Selection Vector $q$, Number of Concepts $k$.
		\Ensure Set of Concepts $C_o$
		\State $C_g \leftarrow  \varnothing $, $C_l \leftarrow  \varnothing$, $C_o \leftarrow  \varnothing$
		\State ${Map}(x, y)\leftarrow G_g(I,M)$
		\State Sort activation scores $v_i$ in ${Map}$ in descending order to obtain $Param_g = \{(v_i, x_i, y_i) \mid i = 1, 2, \ldots, n\}$
		\State $C_g \leftarrow D_{SAM}(I, Param_g, q, k)$	
		\State $S \leftarrow G_l(I, M, k)$
		\State $L_b \leftarrow  \varnothing$
		\For{each superpixel $s_i \in S$}
		    \State Get the bounding box $p_i = (x_{i1}, y_{i1}, x_{i2}, y_{i2})$
		    \State $L_b \gets L_b \cup \{p_i\}$
		\EndFor
		\State  $C_l \leftarrow D_{SAM}(I, L_b, k)$
		\State $C_o \gets C_g \cup  C_l$
		\State \textbf{return} $C_o$
	\end{algorithmic}
\end{algorithm}

\section{Methodology}
\subsection{SAM-based concept discovery for lesions}
As shown in the first row of Fig. \ref{fig:method}., a typical DNN black-box model \(f\) is processed through the $softmax$ layer after receiving an image $I$ $\in$ \(\mathbb{R}^{w\times h \times c}\) to obtain a probability vector of length $k$ of the number of classification categories, where the largest probability corresponds to the classification result. Such a decision-making process is entirely closed-off, and we cannot discern what the input image has undergone within the black-box to arrive at the final classification result. 

Now, we try to use SAM\cite{kirillov2023segment} for concept discovery in the black-box. As a zero-shot segmentation model, given an image \(I\) and either a set of points $L_p$ of length $n$, a set of bounding boxes $L_b$ of length $n$, or no prompt provided (in ``everything" mode), SAM can output a set of $n$ concept masks $C = \{c_1, c_2, ..., c_n\}$, where each $c_i$ may correspond to certain concepts in real-world. In other words, SAM can be seen as having the ability to discover concepts. Therefore, we redefine SAM as a concept discoverer $D_{SAM}$. Given the limited concept discovery capability of the original $D_{SAM}$ on medical images, the $D_{SAM}$ used in this paper is SAMMed2D, which is fine-tuned on a large medical image dataset (see Section 4. B for comparison results between different training versions).

Concept discovery directly using $D_{SAM}$ does not result in meaningful and high-quality concept blocks on ultrasound images. This is because ultrasound images, as a low-cost image modality, contain relatively limited information compared to normal, CT, or MRI images. In contrast, many objects in normal images have specific concepts that are easy to discover. Therefore, we consider the use of attribution methods such as GradCAM \cite{selvaraju2017grad} and LIME \cite{ribeiro2016should} as concept guides to facilitate the discovery of concepts containing specific lesions information. Algorithm \ref{al1} describes the whole process of concept discovery, each concept is saved in the form of a mask.

For pixel-level and superpixel-level attribution methods, we employ different guiding approaches. (1) Concept guide $G_g$: GradCAM\cite{selvaraju2017grad} is a pixel-level attribution method that utilizes gradient information to obtain activation maps of the model for a particular class outcome. Given an image $I$ and a target model $M$, GradCAM generates a heatmap ${Map}(x_i, y_i) = v_i$, where $v_i$ represents the attribution score obtained by GradCAM, and $x_i$, $y_i$ are the coordinates of the corresponding point. The triplets $(v_i, x_i, y_i)$ are sorted in descending order based on the value of $v_i$ to obtain the guiding parameters for GradCAM, denoted as $Param_g = \{(v_i, x_i, y_i)\}$. Based on the selection vector $q$, the corresponding triplets are obtained from $Param_g$ in the specified percentages. These triplets are then input into $D_{SAM}$ to obtain the concept set $C_g$ guided by GradCAM. (2) Concept guide $G_l$: LIME is a perturbation-based attribution method where explanations are presented in the form of superpixels. By using $G_l$, the set of superpixels $S$ can be obtained. For each superpixel $s_i$ in $S$, get the bounding box and put it into $L_b$. Through $L_b$, the concept set $C_l$ guided by LIME can be obtained using $D_{SAM}$. Finally, $C_l$ is combined with $C_g$ to obtain the unprocessed $C_o$.

\subsection{Concepts post-processing and Shapley value calculation}

The concepts mask guided by $G_g$ and $G_l$ may have obvious overlaps, which can lead to unnecessary computational waste and explanation errors in the subsequent process. For each $c_i$ $\in$ $C_o$, if the intersection between any two concepts is more than 90\%, the concept with the smaller number of pixels is discarded until all elements in $C_o$ satisfy this condition. The explanation $E$, which contains the meaning of a lesion, is a subset of $C_o$. As with EAC\cite{sun2024explain}, we use the Shapley value \cite{shapley1953value,lundberg2017unified} for the final explanation. The model's original prediction for the image will serve as the utility function $u(T)$, where for concept $c_i$ its marginal contribution to the model's decision is defined as 
\begin{equation}
\phi_{c_i}(I)=\mathbb{E}\left(u\left(T \cup\left\{c_i\right\}\right)-u(T)\right)
\end{equation}

Since the number of concepts is determined by the hyperparameters, there is no need to use Monte Carlo approximations to calculate Shapley values\cite{sun2024explain}. In addition, avoiding the use of surrogate models\cite{sun2024explain} avoids a decrease in the faithfulness of the explanation. Finally, the optimal explanation is defined as
\begin{equation}
E=\operatorname{argmax}_{E \in C} \phi_E(I)
\end{equation}

\begin{figure}[tbp]
  \centering
  \includegraphics[width=0.45\textwidth]{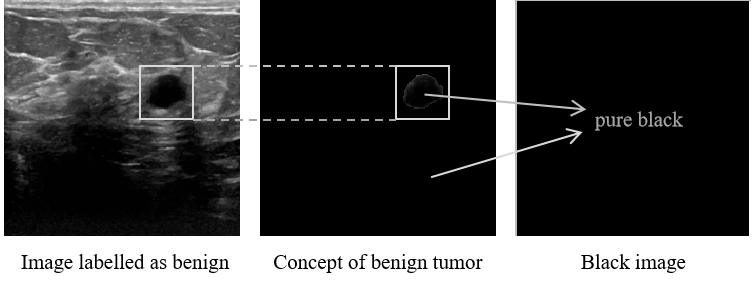}
  \caption{The Shapley value calculated when a black image is diagnosed as benign with a 96\% probability is not faithful.}
  \label{fig:Shapley}
\end{figure}

When combining $T$, we observe that the concept of benign tumors is intuitively very similar to the black pixel blocks used for masking, as shown in Fig. \ref{fig:Shapley}. Black image have a 96\% probability of being diagnosed as benign when the model does not suffer from overfitting problems. Therefore, there is no guarantee that the calculated Shapley Value faithfully reflects the marginal contribution of $C_i$ when black masking is used. For this reason, we decided to replace the black pixels with randomly selected pixels from a normal distribution whose mean and variance matched those of the source dataset.
\section{Experiments and analysis}
\subsection{Experiment setting}
\subsubsection{Datasets}
Breast Ultrasound Image (BUSI)\cite{al2020dataset} is a classification and segmentation dataset containing breast ultrasound images. For classification labeling, there are three categories, normal, benign, and malignant, and the sample size for each category is shown in Table \ref{tbl1}.

Fine-grained Ultrasound Images Of The Breast (FG-US-B) is a private fine-grained classification dataset of breast ultrasound images, and the classification labels are all based on the pathological results of the puncture procedure. The total number of images is 1016 and the sample size for each category is shown in Table \ref{tbl1}. 

\begin{figure}[tbp]
    \centering
    \subfloat[Normal]{\includegraphics[width=0.10\textwidth]{BUSI-normal}}
    \subfloat[Benign]{\includegraphics[width=0.10\textwidth]{BUSI-benign}}
    \subfloat[Malignant]{\includegraphics[width=0.10\textwidth]{BUSI-malignant}}
    \subfloat[DCIS]{\includegraphics[width=0.10\textwidth]{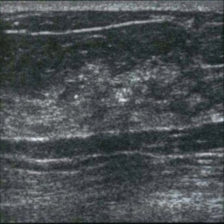}}
    \subfloat[Cyst]{\includegraphics[width=0.10\textwidth]{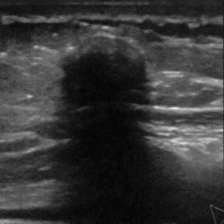}}\\
    \subfloat[IP]{\includegraphics[width=0.10\textwidth]{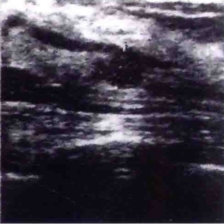}}
    \subfloat[IDC]{\includegraphics[width=0.10\textwidth]{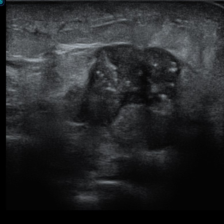}}
    \subfloat[Mastitis]{\includegraphics[width=0.10\textwidth]{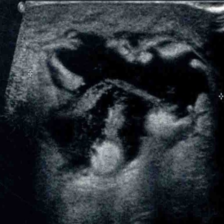}}
    \subfloat[MP]{\includegraphics[width=0.10\textwidth]{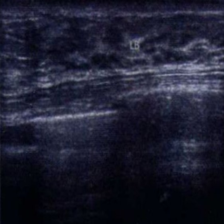}}
    \subfloat[FN]{\includegraphics[width=0.10\textwidth]{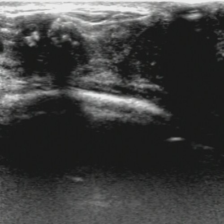}}
    \caption{Examples of datasets.}
    \label{fig3}
\end{figure}

The sample illustrations of the datasets used are shown in Fig. \ref{fig3}. The FG-US-B dataset was collected from the ultrasound department of the Affiliated Hospital of Southwest Jiaotong University (Chengdu Third People's Hospital).
\subsubsection{Implementation details}

The main research aim of this paper is a new framework for explainability rather than the structure of a specific model. In fact, our framework is model-agnostic. We use ResNet50 \cite{he2016deep} as the target model, where the images are resized to (224, 224) before being input into the model. During training, we apply random horizontal flipping and calculate normalization coefficients separately for the two datasets. The optimization process uses decoupled weight decay (AdamW) \cite{loshchilov2017decoupled}, with an initial learning rate of $5 \times 10^{-3}$, gradually decreasing to $5 \times 10^{-5}$ during fine-tuning. Each iteration uses a batch size of 8 samples. The implementation is carried out using the PyTorch framework (version 2.1.2 and CUDA 12.3) \cite{paszke2017automatic}, and all experiments are performed on a GeForce RTX 3090 GPU. The classification accuracy of the target model on the test set after training is shown in Table \ref{tbl1}.

\begin{table}[tbp]
    \caption{Amount of samples in the datasets and classification accuracy of ResNet50 \cite{he2016deep}.}
    \label{tbl1}
    \centering
    \begin{tabular}{|c|c|>{\centering\arraybackslash}p{1.0cm}|>{\centering\arraybackslash}p{1.0cm}|}
        \hline
        \textbf{Dataset} & Class & Amount & Accuracy \\
        \hline
        \multirow{7}{*}{FG-US-B} & Mastopathy (Ma) & 78 & 30.76 \\
        \cline{2-4}
        & Mastitis & 126 & 53.65 \\
        \cline{2-4}
        & Fibroadenoma (FN) & 258 & 94.44 \\
        \cline{2-4}
        & Intraductal papilloma (IP) & 133 & 67.44 \\
        \cline{2-4}
        & Cyst & 104 & 60.71 \\
        \cline{2-4}
        & Ductal carcinoma in situ (DCIS) & 93 & 60.00 \\
        \cline{2-4}
        & Invasive ductal carcinoma (IDC) & 224 & 80.00 \\
        \hline
        \multirow{3}{*}{BUSI} & Normal & 133 & 75.00 \\
        \cline{2-4}
        & Benign & 437 & 89.39 \\
        \cline{2-4}
        & Malignant & 210 & 85.71 \\
        \hline
    \end{tabular}
\end{table}

\subsection{Evaluation metrics}
We evaluate the methods in terms of both understandability and faithfulness. Faithfulness aims to respond to whether the explanation is the same as the model's decision-making process. Intuitively, a good explanation should be human-understandable \cite{rasheed2022explainable}. Based on the principle of human-in-the-loop\cite{lage2018human}, we invite sonographers to perform an evaluation of understandability, see Section 4.E.

The $Insertion$ and $Deletion$\cite{petsiuk2018rise} are popular faithfulness metrics. The $Insertion$ measures the importance of the superpixels in constructing the whole image, while the $Deletion$ quantifies the extent to which the probability decreases as important superpixels are gradually removed from the image.

\begin{figure}[tbp]
    \centering
    \subfloat[$\mathrm{E}_1$ (EAC)]{\includegraphics[width=0.16\textwidth]{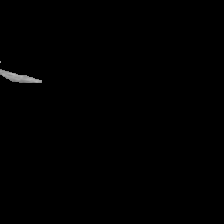}\label{fig:sub1}}\hfill
    \subfloat[$\mathrm{E}_3$ (EAC)]{\includegraphics[width=0.16\textwidth]{EAC_2}\label{fig:sub3}}\hfill
    \subfloat[$Insertion$ (EAC)]{\includegraphics[width=0.16\textwidth]{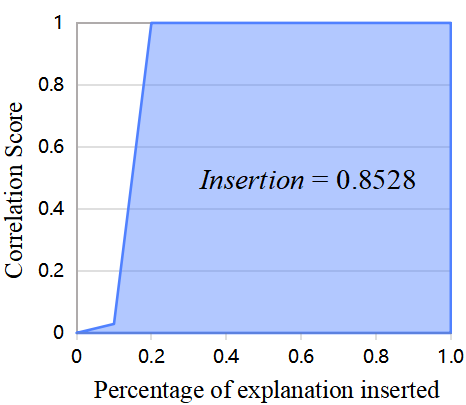}\label{fig:sub4}}\hfill
    \subfloat[$\mathrm{E}_1$ (LIME)]{\includegraphics[width=0.16\textwidth]{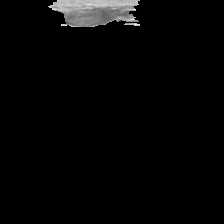}\label{fig:sub6}}\hfill
    \subfloat[$\mathrm{E}_3$ (LIME)]{\includegraphics[width=0.16\textwidth]{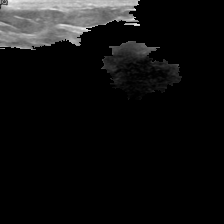}\label{fig:sub8}}\hfill
    \subfloat[$Insertion$ (LIME)]{\includegraphics[width=0.16\textwidth]{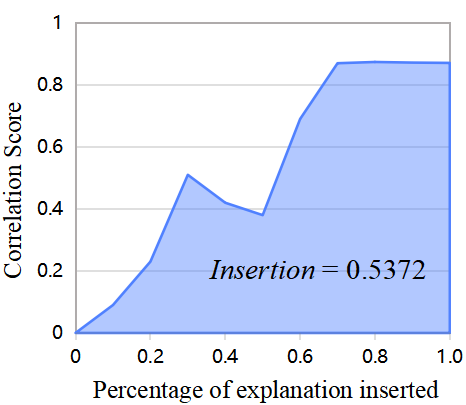}\label{fig:sub9}}\hfill
    \caption{The $Insertion$ curve responds to the extent to which the confidence of the model predictions correlates with the original predictions as the percentage of explanation inserted increases. The EAC generates some meaningless explanations, but those explanations receive high evaluation results (first row). In contrast, LIME generates explanations that do not match certain concepts, but are not as meaningless as EAC (second row). $\mathrm{E}_i$ indicates that this is the $i$-th explanation accumulation image produced in the computation of $Insertion$.}

    \label{fig4}
\end{figure}

We found that $Insertion$ and $Deletion$ could not reasonably evaluate the faithfulness of EAC\cite{sun2024explain} on breast ultrasound images due to the inability to determine the number and size of concepts discovered. Although EAC performs well on $Insertion$ (see the first row of Fig. \ref{fig4}), its evaluation is not reasonable. This is because $\mathrm{E}_3$ (EAC) shows almost the whole image, resulting in the best evaluation. In contrast, LIME\cite{ribeiro2016should} ranks lower, but not so low as to be meaningless. We consider the size of the visual explanation to be a key factor in evaluating the quality of the explanation. If the explanation is too large, taking up the same proportion of the whole image, then the explanation becomes meaningless. Evaluation on $Deletion$ has similar results.

For the above reasons, we modify these two faithfulness metrics. Define $P_o$ as the number of pixels in the input image, and $P_s$ denotes the average number of pixels in a set of explanation accumulation images. The weighted evaluation metrics $Insertion_w$ and $Deletion_w$ are defined as
\begin{equation}
Insertion_w =\left(1-\frac{P_s}{P_o}\right) \times Insertion
\end{equation}
\begin{equation}
Deletion_w =\left(1-\frac{P_s}{P_o}\right)\times(1-Deletion)
\end{equation}

Considering both $Insertion_w$ and $Deletion_w$, the evaluation metric $Effect\ Score$ is defined as
\begin{equation}
Effect\ Score =\frac{Insertion_w+Deletion_w}{2}
\end{equation}

In the next experiments, in addition to $Insertion$ and $Deletion$, we use the $Effect\ Score$ to evaluate the faithfulness of methods.

\subsection{Comparison of SAMs trained on different data}
The concept discovery capability of SAM is crucial to our framework. We compare the faithfulness of different $D_{SAM}$, including SAMs with different encoders\cite{dosovitskiy2020image} and SAMs fine-tuned on different medical image datasets (SAMMed2D \cite{cheng2023sam} and MedSAM \cite{ma2024segment}). To evaluate $D_{SAM}$ reasonably, we exclude $F_l$ and $F_g$. The experimental results are summarized in Table \ref{tab2}.

\begin{table*}[htbp]
\caption{Comparing the faithfulness of $D_{SAM}$ of different encoders and $D_{SAM}$ fine-tuned on different data, $\uparrow$ and $\downarrow$ indicate that higher is better and lower is better, respectively.}
\label{tab2}
\renewcommand\arraystretch{1.1} % Adjust row height
\centering
\begin{tabular}{|c|*{6}{c|}}
\hline
\multirow{2}{*}{$D_{SAM}$} & \multicolumn{3}{c|}{BUSI} & \multicolumn{3}{c|}{FG-US-B} \\
\cline{2-7}
& $Insertion$ $\uparrow$ & $Deletion$ $\downarrow$ & $Effect\ Score$ $\uparrow$ &$Insertion$ $\uparrow$ & $Deletion$ $\downarrow$ & $Effect\ Score$ $\uparrow$ \\
\hline
SAM(ViT-B) & 0.7540 & 0.5885 & 0.3293 & 0.4885 & 0.4817 & 0.1865 \\
\hline
SAM(ViT-L) & 0.7732 & 0.5803 & 0.2772 & \textbf{0.6473} & 0.3299 & 0.1793 \\
\hline
SAM(ViT-H) & \textbf{0.8675} & \textbf{0.5497} & 0.1768 & 0.6470 & 0.3123 & 0.1762 \\
\hline
MedSAM \cite{ma2024segment} & 0.5774 & 0.6000 & 0.3612 & 0.1214 & \textbf{0.1702} & 0.3247 \\
\hline
SAMMed2d\cite{cheng2023sam} & 0.5954 & 0.5525 & \textbf{0.3785} & 0.1326 & 0.2035 & \textbf{0.3365} \\
\hline
\end{tabular}
\end{table*}

\begin{table*}[htbp]
\caption{Comparing the faithfulness of different methods, $\uparrow$ and $\downarrow$ indicate that higher is better and lower is better, respectively.}
\label{tab3}
\renewcommand\arraystretch{1.1} % Adjust row height
\centering
\begin{tabular}{|c| *{6}{c|}}
\hline
\multirow{2}{*}{Method} & \multicolumn{3}{c|}{BUSI} & \multicolumn{3}{c|}{FG-US-B} \\
\cline{2-7}
& $Insertion$ $\uparrow$ & $Deletion$ $\downarrow$ & $Effect\ Score$ $\uparrow$ & $Insertion$ $\uparrow$ & $Deletion$ $\downarrow$ & $Effect\ Score$ $\uparrow$ \\
\hline
EAC\cite{sun2024explain}  & \textbf{0.8675} & 0.5497 & 0.1768 & \textbf{0.6470} & 0.3123 & 0.1762 \\
\hline
LIME\cite{ribeiro2016should} & 0.6205 & 0.4706 & 0.5073 & 0.1972 & 0.2344 & 0.3131 \\
\hline
IntGrad\cite{sundararajan2017axiomatic} & 0.5964 & 0.4779 & 0.3647 & 0.1849 & 0.2252 & 0.3118 \\
\hline
GradSHAP\cite{lundberg2017unified} & 0.6035 & 0.4771 & 0.3674 & 0.2144 & 0.2818 & 0.3029 \\
\hline
DeepLIFT\cite{shrikumar2017learning} & 0.6510 & 0.5045 & 0.3733 & 0.2938 & 0.2309 & 0.3446 \\
\hline
KernelSHAP\cite{lundberg2017unified} & 0.4797 & 0.5763 & 0.2945 & 0.1778 & 0.2807 & 0.2915 \\
\hline
GradCAM\cite{selvaraju2017grad} & 0.7175 & \textbf{0.4287} & 0.4172 & 0.1624 & 0.1936 & 0.3315 \\
\hline
LCE & \textbf{0.7585} & 0.4750 & \textbf{0.5572} & 0.4477 & \textbf{0.1889} & \textbf{0.4947}\\
\hline
\end{tabular}
\end{table*}
\setlength{\tabcolsep}{1.5pt}
\begin{figure*}[htbp]
    \centering

    % Column titles
    \begin{tabular}{ccccccccc}
        \multicolumn{1}{c}{Input} & 
        \multicolumn{1}{c}{LCE} & 
        \multicolumn{1}{c}{EAC} & 
        \multicolumn{1}{c}{LIME} & 
        \multicolumn{1}{c}{IntGrad} & 
        \multicolumn{1}{c}{GradSHAP} & 
        \multicolumn{1}{c}{DeepLIFT} & 
        \multicolumn{1}{c}{KernalSHAP} & 
        \multicolumn{1}{c}{GradCAM} \\

        % First row of images
        \subfloat{\includegraphics[width=0.106\textwidth]{Fig5/1}} &
        \subfloat{\includegraphics[width=0.106\textwidth]{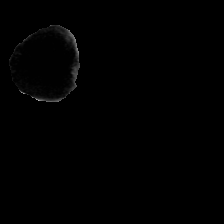}} &
        \subfloat{\includegraphics[width=0.106\textwidth]{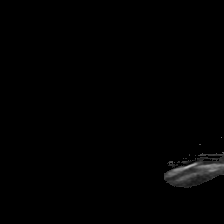}} &
        \subfloat{\includegraphics[width=0.106\textwidth]{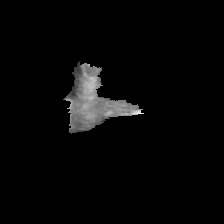}} &
        \subfloat{\includegraphics[width=0.106\textwidth]{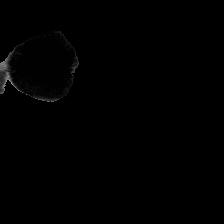}} &
        \subfloat{\includegraphics[width=0.106\textwidth]{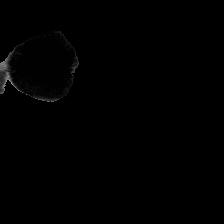}} &
        \subfloat{\includegraphics[width=0.106\textwidth]{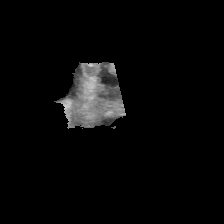}} &
        \subfloat{\includegraphics[width=0.106\textwidth]{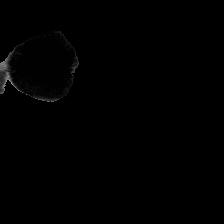}} &
        \subfloat{\includegraphics[width=0.106\textwidth]{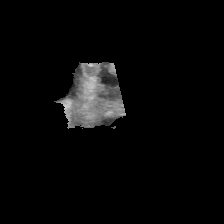}} \\

        % Second row of images
        \subfloat{\includegraphics[width=0.106\textwidth]{Fig5/2-1}} &
        \subfloat{\includegraphics[width=0.106\textwidth]{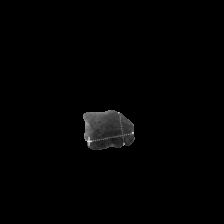}} &
        \subfloat{\includegraphics[width=0.106\textwidth]{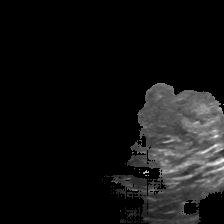}} &
        \subfloat{\includegraphics[width=0.106\textwidth]{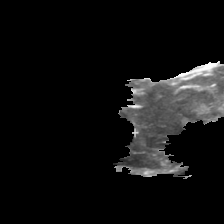}} &
        \subfloat{\includegraphics[width=0.106\textwidth]{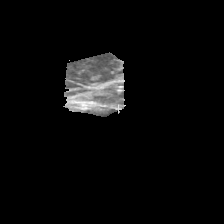}} &
        \subfloat{\includegraphics[width=0.106\textwidth]{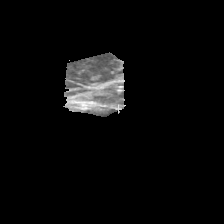}} &
        \subfloat{\includegraphics[width=0.106\textwidth]{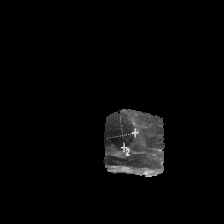}} &
        \subfloat{\includegraphics[width=0.106\textwidth]{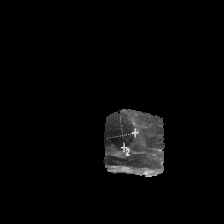}} &
        \subfloat{\includegraphics[width=0.106\textwidth]{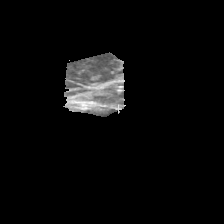}} \\

        % Third row of images
        \subfloat{\includegraphics[width=0.106\textwidth]{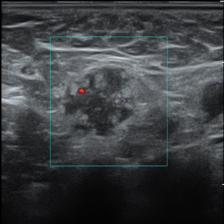}} &
        \subfloat{\includegraphics[width=0.106\textwidth]{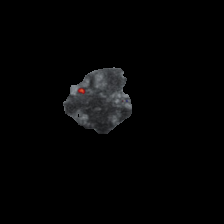}} &
        \subfloat{\includegraphics[width=0.106\textwidth]{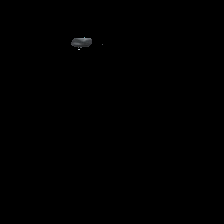}} &
        \subfloat{\includegraphics[width=0.106\textwidth]{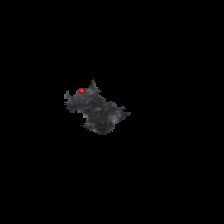}} &
        \subfloat{\includegraphics[width=0.106\textwidth]{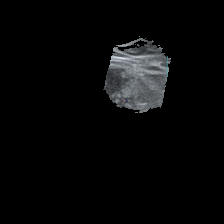}} &
        \subfloat{\includegraphics[width=0.106\textwidth]{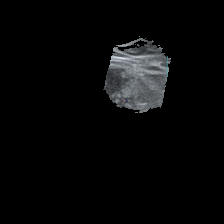}} &
        \subfloat{\includegraphics[width=0.106\textwidth]{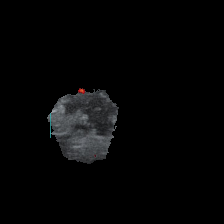}} &
        \subfloat{\includegraphics[width=0.106\textwidth]{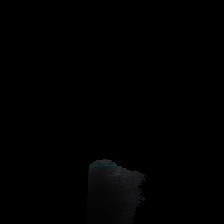}} &
        \subfloat{\includegraphics[width=0.106\textwidth]{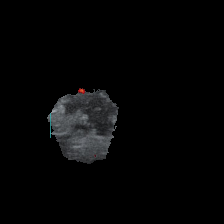}} \\

        % Fourth row of images
        \subfloat{\includegraphics[width=0.106\textwidth]{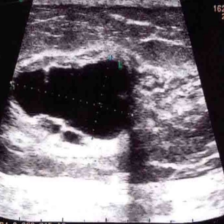}} &
        \subfloat{\includegraphics[width=0.106\textwidth]{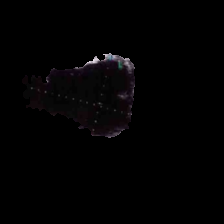}} &
        \subfloat{\includegraphics[width=0.106\textwidth]{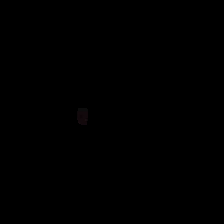}} &
        \subfloat{\includegraphics[width=0.106\textwidth]{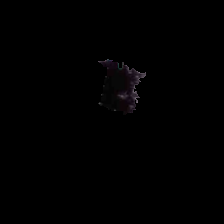}} &
        \subfloat{\includegraphics[width=0.106\textwidth]{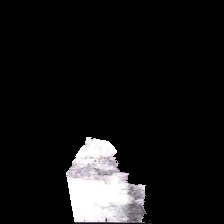}} &
        \subfloat{\includegraphics[width=0.106\textwidth]{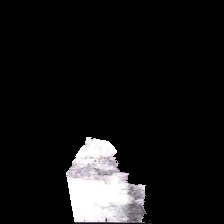}} &
        \subfloat{\includegraphics[width=0.106\textwidth]{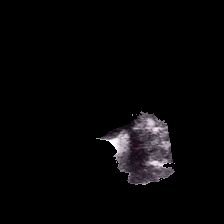}} &
        \subfloat{\includegraphics[width=0.106\textwidth]{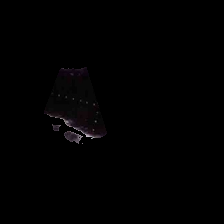}} &
        \subfloat{\includegraphics[width=0.106\textwidth]{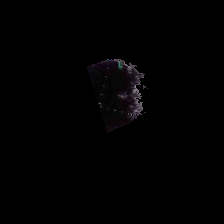}} \\

    \end{tabular}
  \caption{Sample explanations generated by LCE, EAC\cite{sun2024explain} and six baseline methods\cite{ribeiro2016should,sundararajan2017axiomatic,lundberg2017unified,shrikumar2017learning,selvaraju2017grad}.}
   \label{fig:comparison}
\end{figure*}

\begin{table*}[h]
\caption{Understandability is evaluated by five professional sonographers, and the table shows the average number of times the explanation generated by each method is considered to be the best explanation.}
\label{tab4}
\centering
\renewcommand\arraystretch{1.1} % Adjust row height
\begin{tabular}{|c|*{7}{>{\centering\arraybackslash}p{1.5cm}|}}
\hline
DataSet & LCE & LIME & IntGrad &
GradSHAP & DeepLIFT & KernelSHAP &GradCAM \\
\hline
BUSI      & \textbf{64} & 30 & 54 & 56 & 45 & 28 & 45 \\
\hline
FG-US-B   & \textbf{52} & 40 & 34 & 32 & 30 & 29 & 38 \\
\hline
\end{tabular}
\end{table*}

\begin{table*}[h]
    \centering
    \caption{Explanations generated by LCE for models trained on data at different fine-grained levels.}
    \label{tab:finegrain}
    \renewcommand\arraystretch{1.1} % Adjust row height
    \begin{tabular}{|>{\centering\arraybackslash}m{1.9cm}|>{\centering\arraybackslash}m{0.12\textwidth}|>{\centering\arraybackslash}m{0.12\textwidth}|>{\centering\arraybackslash}m{0.12\textwidth}|>{\centering\arraybackslash}m{0.12\textwidth}|>{\centering\arraybackslash}m{0.12\textwidth}|>{\centering\arraybackslash}m{0.12\textwidth}|>{\centering\arraybackslash}m{0.12\textwidth}|}
        \hline
        \multirow{4}{=}{\centering Original image\\ and explanations}  & \raisebox{-.5\height}{\includegraphics[width=0.12\textwidth]{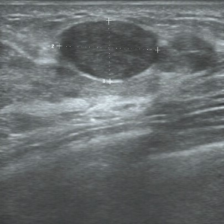}} & \raisebox{-.5\height}{\includegraphics[width=0.12\textwidth]{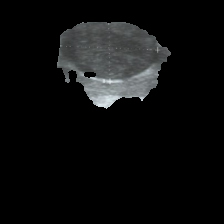}} & \raisebox{-.5\height}{\includegraphics[width=0.12\textwidth]{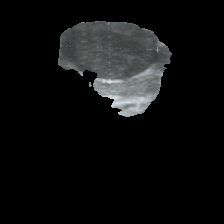}} & \raisebox{-.5\height}{\includegraphics[width=0.12\textwidth]{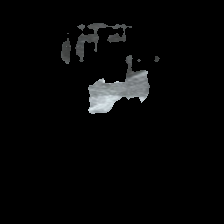}} & \raisebox{-.5\height}{\includegraphics[width=0.12\textwidth]{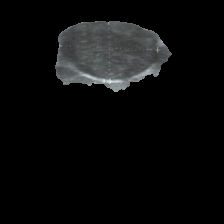}} & \raisebox{-.5\height}{\includegraphics[width=0.12\textwidth]{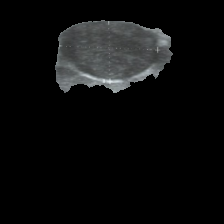}} & \raisebox{-.5\height}{\includegraphics[width=0.12\textwidth]{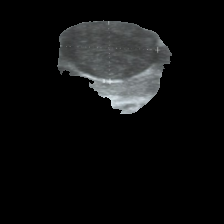}} \\
        \hline
        Number of categories & - & 2 & 3 & 4 & 5 & 6 & 7 \\
        \hline
        Successful diagnosis & - & Yes & Yes & No & Yes & Yes & Yes \\
        \hline
        $Effect\ Score$ & - & 0.4429 & 0.4816 & 0.4539 & 0.5013 & 0.4926 & 0.4823 \\
        \hline
    \end{tabular}
\end{table*}

%\begin{figure*}[tbp]
%  \centering
%  \includegraphics[width=1.0\textwidth]{comparison}
%  \caption{Sample explanations generated by LCE, EAC\cite{sun2024explain} and six baseline methods\cite{ribeiro2016should,sundararajan2017axiomatic,lundberg2017unified,shrikumar2017learning,selvaraju2017grad}.}
%  \label{fig:comparison}
%\end{figure*}
The original SAM (ViT-H version) has not been trained on a large amount of medical image data and has the best $Insertion$ and $Deletion$. As discussed in Section 4.B, this is due to the fact that its explanation is the image itself. After re-evaluation using the $Effect\ Score$, these highly unreliable explanations are distinguished from the others. SAMMed2D is considered to be the best $D_{SAM}$ due to its fine-tuning on the largest medical image dataset to date. To maintain the best performance, LCE uses SAMMed2D as $D_{SAM}$.
\subsection{Comparison of different explainability methods}
In addition to the EAC, which is the most similar to the LCE, we select six commonly-used baseline methods for comparison. These methods are all implemented by captum\cite{kokhlikyan2020captum}. We first generate concepts from the input images using superpixels\cite{achanta2012slic} and then calculate the faithfulness of these concepts. The comparison results are shown in Table \ref{tab3}.
Among all the methods, LCE has the highest $Effect\ Score$. On the BUSI dataset, it has an $Effect\ Score$ of 0.5572, which is 5\% higher than the second-ranked LIME. Similarly, on the FG-US-B dataset, it is 15\% higher than the second-ranked method. This suggests that LCE consistently produces more faithful breast ultrasound explanations than the baseline method.

We present several explanations generated by different methods in Fig. \ref{fig:comparison}. Compared to other methods, LCE always gives understandable explanations that are consistent with the lesion concept. For example, in the third row, EAC generates an explanation of very small super-pixels, and this explanation makes no sense. In contrast, other baseline methods generate explanations that are related to the tumor concept but are less specific than LCE. While maintaining a certain level of faithfulness, after concept guidance, LCE can discover meaningful concepts (e.g. the concept of a tumor) from low-cost ultrasound images instead of presenting whole images or meaningless almost completely black images.

%
%\begin{figure*}[htbp]
%  \centering
%  \includegraphics[width=1.0\textwidth]{finegrain}
%  \caption{Explanations of LCE for models trained at different fine-grained levels.}
%  \label{fig:finegrain}
%\end{figure*}

\subsection{Understandability evaluation from sonographers}

Feature maps can also be used as explanations, but such explanations are not valid because humans cannot understand them. Many methods evaluate understandability through human-centered experiments \cite{colin2022cannot, ribeiro2016should,sun2024explain}. In this paper, we select samples of benign, fibroadenoma, and mastitis for understandability evaluation.

We invite five experienced sonographers to evaluate the baseline methods and LCE. We provide each sonographer with the raw image, the model prediction, the ground truth, and seven explanations for each image. The order in which each explanation is presented is randomized, so the sonographers are unaware of which method each explanation comes from. They are required to rank the regions included in the seven explanations, with the explanations that contributed most to the diagnostic results and are easier to understand being ranked highest. The evaluation is completed when the sonographers are fully rested. The average evaluation of the five sonographers is shown in Table \ref{tab4}. Based on the sonographers' ranking results, LCE clearly outperforms the other baseline methods. The majority of the explanations provided by the LCE are accepted by the sonographers. In particular, this shows that LCE contributes to the realization of AI that can be trusted by medical professionals.

\subsection{Explanations at different diagnostic fine-grained levels}
Fine-grained classifications are those with subtle differences within categories, and existing studies tend to classify breast tumors as benign or malignant only on a coarse-grained basis\cite{shia2021classification}. However, coarse-grained classifications serve a weak auxiliary role for sonographers, who can easily derive a diagnosis from several guidelines (e.g., TI-RADS \cite{tessler2017acr}).

To test the performance of LCE on more significant fine-grained diagnosis, we create subsets of six different fine-grained levels by changing the number of categories using the FG-US-B dataset, and Table \ref{tab:finegrain} shows the differences in the explanations generated by LCE as the change of fine-grained levels. When the model's diagnosis is incorrect, we can also find that its explanation becomes ambiguous, reflecting to some extent the reason for the incorrect diagnosis (e.g. failure to correctly identify the lesion).
\section{Conclusion}
In this paper, we integrate the attribution method with concept-based method by proposing the Lesion Concept Explainer (LCE) framework, which uses SAM fine-tuned on a large number of medical images for lesion concept discovery, to explain the decisions of DNNs. To address the shortcomings of the popular faithfulness evaluation metrics $Insertion$ and $Deletion$, we modify them by using the size of the explanation as a weight and propose a new metric, $Effect\ Score$. We implement LCE on ResNet50 models trained on two breast ultrasound datasets (public and private). Compared to other popular methods, LCE is more effective in explaining breast ultrasound DNNs and performs best in the evaluation of objective faithfulness and human-centered understandability. Finally, we experiment with LCE at different fine-grained levels and show that it always faithfully provides understandable explanations. More importantly, we demonstrate that LCE can contribute to the development of medical AI by making DNNs in the medical domain more trustworthy.

\bibliographystyle{IEEEtran}
\bibliography{IEEEabrv,reference}

\end{document}